\documentclass[
]{ceurart}

\usepackage{listings}
\usepackage{tcolorbox}
\begin{document}

\copyrightyear{2023}
\copyrightclause{Copyright for this paper by its authors.
  Use permitted under Creative Commons License Attribution 4.0
  International (CC BY 4.0).}

\conference{Forum for Information Retrieval Evaluation, December 15-18, 2023, India}

\title{Overview of the HASOC Subtrack at FIRE 2023: Identification of Tokens Contributing to Explicit Hate in English by Span Detection}


\author[1]{Sarah Masud}[%
email=sarahm@iiitd.ac.in
]
\cormark[1]
\address[1]{Indraprastha Institute of Information Technology, Delhi}

\author[1]{Mohammad Aflah Khan}[%
email=aflah20082@iiitd.ac.in
]

\author[1]{Md. Shad Akhtar}[%
email=shad.akhtar@iiitd.ac.in
]
\author[2]{Tanmoy Chakraborty}[%
email=tanchak@iitd.ac.in
]
\address[2]{Indian Institute of Technology, Delhi}

\cortext[1]{Corresponding author.}

\begin{abstract}
As hate speech continues to proliferate on the web, it is becoming increasingly important to develop computational methods to mitigate it. Reactively, using black-box models to identify hateful content can perplex users as to why their posts were automatically flagged as hateful. On the other hand, proactive mitigation can be achieved by suggesting rephrasing before a post is made public. However, both mitigation techniques require information about which part of a post contains the hateful aspect, i.e., what spans within a text are responsible for conveying hate. Better detection of such spans can significantly reduce explicitly hateful content on the web. To further contribute to this research area, we organized \textit{HateNorm} at HASOC-FIRE 2023, focusing on explicit span detection in English Tweets\footnote{\color{red}{\noindent\textbf{Disclaimer:} The paper contains samples of hate speech, which are only included for contextual understanding. We do not support these views.}}. A total of $12$ teams participated in the competition, with the highest macro-F1 observed at $0.58$.
\end{abstract}

\begin{keywords}
  Hate Span \sep
  Explicit Hate \sep
  English Tweet \sep
  HASOC'23
\end{keywords}

\maketitle

\section{Introduction}
Hate speech is a challenging social issue given its subjective nature: what is hateful changes with time, geography, and cultural context. United Nations defines hate speech\footnote{\url{https://www.un.org/en/hate-speech/understanding-hate-speech/what-is-hate-speech}} as ``any form of communication that uses pejorative or discriminatory language with reference to a person or a group based on who they are." Further, hate speech has real-world implications; not only do real-world biases drive up online hate speech, but online hate speech can lead to an increase in hate crimes in the offline world. To reduce the burden on the volume and velocity of hateful content accessed by content moderators, analyzing, mitigating, and countering hateful content via computational methods is binding. While computer-aided can help perform the first level of mitigation, human involvement in subjective matters like hate speech is critical and compulsory for improving social systems in the real world. To aid the systems in better detection of hateful content, one can look into developing systems that can capture and attend to the hateful spans \cite{pavlopoulos-etal-2021-semeval} within a sentence. Span detection can help develop a sense of rationale, act as a tool for post hoc analysis, and improve the retrival of critical facts in claim verification \cite{sundriyal-etal-2022-empowering}. 

\begin{table*}[!t]
    \centering
    \caption{A few examples of hateful posts of varying degrees from the dataset curated by \citet{hatenorm} and their corresponding hateful spans marked in red.}
    \label{tab:motivation}
    \resizebox{\textwidth}{!}
    {
    \begin{tabular}{c|p{40em}}
    \hline
    \textbf{Sample \#} & \bf Text \\ \hline \hline
    1 & Women ... Can't live with them ... {\textcolor{red}{Can't shoot them}}\\ \hline
    2 & kathy griffin is the ultimate liberal attack somebody and suffer consequences and 5 \& she is the victim \textcolor{red}{c**t needs to have some dick forced up her dy**} ass barron trumps \\ \hline
    \end{tabular}}
\end{table*}

\textbf{Shared Task Objective.}  A hate span is a set of continuous tokens that, in tandem, communicate the explicit hatefulness in a sentence. Table \ref{tab:motivation} provides some examples of harmful social media posts marked for hateful fragments. For instance, in the first sentence of Table \ref{tab:motivation}, \textit{``Women ... Can’t live with them ... \{\textcolor{red}{Can’t shoot them}\}"}, the portion highlighted in red will be considered as a hateful span. Formally, \textit{given a hate sample, tokenized as $t = \langle w_1, w_2, \dots, w_n\rangle$, the hate span identification task looks for a sequence of hateful tokens, $\langle w_i,.., w_{i+l} \rangle$} \cite{hatenorm}.

\begin{tcolorbox}[width=\textwidth]    
\textbf{Problem Definition:} \textit{Given a hateful text, identify those specific fragments within the sentence that are hateful. This is a sequence tagging task where the aim is to label each word as either belonging to the hate span or not.}
\end{tcolorbox}

\textbf{Share Task Details.} Through the \textit{HateNorm} shared task part of HASOC-FIRE 2023\footnote{\url{https://hasocfire.github.io/hasoc/2023/}}, we aimed at engaging the broader research community in understanding span detection techniques and contributing towards the extraction of spans inside a hateful text. In this task, we repurpose a part of the publicly available Hate Normalization dataset \cite{hatenorm}, with each data point containing at least one hate span. The competition ran for a month, from July 13, 2023, to August 16, 2023, PST. Hosted on Kaggle, the task received $72$ submissions from $12$  teams.

\textbf{Observations}
As opposed to a single label per input sentence in a general NLP classification setup, for \textit{HateNorm}, we had a label per token of the sentence \cite{NIPS2017_3f5ee243}. Given the sequential nature of the output, we observed initial hesitation among participants in working with the dataset. However, their engagement improved once a starter kit/codebook was shared. We also observed that among the submissions that submitted a demo paper, the base architecture was more than just a large language model (LLM)-based classifier. There was a mixed usage of both LLM and Bi-directional LSTMs.
Further, we noticed that half of the teams did not apply a CRF layer to capture the sequential encoding of the target label but instead relied on LLM's ability to capture context while making predictions for individual tags. The winning team `FiRC-NLP` with $8$ submissions, obtained macro-F1 scores of $0.53$ and $0.58$ on the public and private leaderboards, respectively. While this beats the start-kit scores of $0.34$, it is comparable to the  SpanBERT-BiLSTM-CRF model from \citet{hatenorm}, which also reported a macro-F1 of $0.58$. More work is needed to bring mainstream attention beyond a text classification of hatefulness to detecting spans. Shared task venues like HASOC and SemEval are the steps in the right direction.

\section{Related Work}
Owing to the relevance and need for computational methods to tackle hate speech, we now have a plethora of datasets \cite{waseem-hovy-2016-hateful,davidson2017automated,10.1145/3580305.3599896} and techniques \cite{schmidt-wiegand-2017-survey, founta-specia-2021-survey} exploring the same. Regarding explicit hate speech, hate lexicons \cite{10.1145/3511047.3537688, stamou-etal-2022-cleansing} have been explored. Auxiliary tasks such as hate normalization \cite{hatenorm,pavlopoulos-etal-2022-detection,agarwal2023haterephrase} and rationale prediction \cite{mathew2021hatexplain} underpinned by the presence or absence of hateful phrases in a sentence led to the foray of hate span detection. In English, the task has been explored from the point of view of detecting toxic and offensive spans \cite {pavlopoulos-etal-2021-semeval,mathew2021hatexplain,hatenorm,pavlopoulos-etal-2022-detection}. In low-resource settings, the span detection has been explored for Vietnamese \cite{hoang-etal-2023-vihos}. Via the MUDES model, \citet{ranasinghe-zampieri-2021-mudes} explored the cross-lingual applicability of hateful span detection when trained on English span datasets. In the multimodal aspect, video frames that conveyed hatefulness were employed as hate spans \cite{Das_Raj_Saha_Mathew_Gupta_Mukherjee_2023}. Another work detected phrases and sentences in long articles that contribute to hate \cite{zhou_caines_pete_hutchings_2023}. In other areas of social computing, span detection has been explored under the English \cite{sundriyal-etal-2022-empowering} and multilingual \cite{mittal2023mcsi} factual claim detection settings. Meanwhile, detecting tokens a model pays attention to while labeling a sample as hateful has been employed in posthoc explanations \cite{kennedy-etal-2020-contextualizing}.

\begin{table}[ht]
    \centering
    \caption{Examples of hateful posts from \citet{hatenorm} dataset and their corresponding \textit{BIO} tags depicting harmful spans.}
    \label{tab:bio_samples}
    \resizebox{0.75\columnwidth}{!}
    {
    \begin{tabular}{p{13em}|p{8em}}
    \hline
         \textbf{Text} & \textbf{Span} \\         
         \hline
         \hline
        lol what a stupid k*k* & \{O, O, O, B, I\}  \\ \hline
        @user text me fa**ot. & \{O, O, O, B\}  \\ \hline
        sad to say but I do not trust shit I know how bi****s operate & \{O, O, O, O, O, B, I, I, I, O, O, O, B, O\} \\ 
        \hline
    \end{tabular}}
\end{table}

\section{Dataset}
\begin{figure}
\centering
\includegraphics[width=0.75\textwidth]{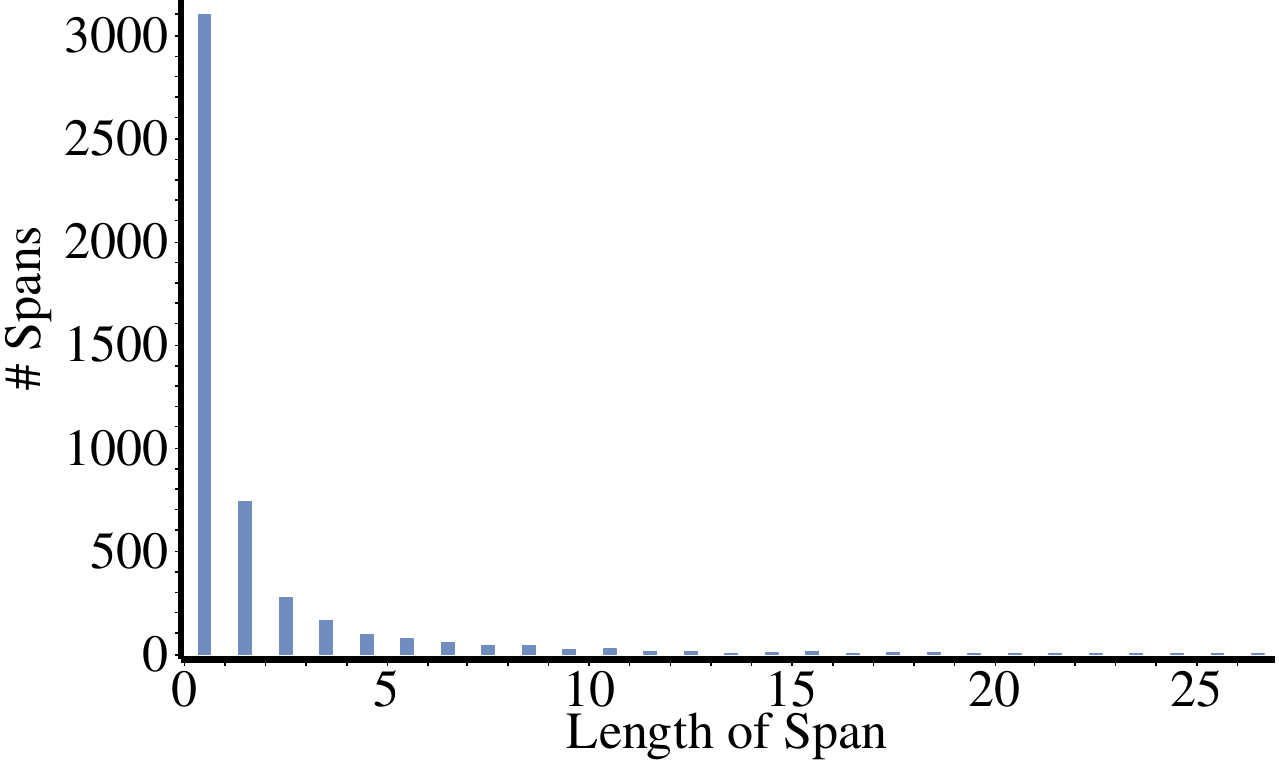}
    \caption{The frequency plot of length of spans in train set. Most samples have only 1 word spans, and majority have spans of length $\leq 5$.}
    \label{fig:enter-label}
\end{figure}

This task employed the existing dataset from \citet{hatenorm} curated initially for $3$ processes -- hate intensity prediction, hate span prediction, and hate normalization generation. We employ only the subset of samples labeled for hate span prediction for hosting \textit{HateNorm}. This led to a dataset with $3027$ explicitly hateful sentences marked with hate spans. As outlined in Table \ref{tab:bio_samples}, the spans are tagged via the BIO notation, marking the beginning and inclusion of span tokens as othering, marking the exclusion. Note that a single token can be a span with a corresponding `B' tag. Meanwhile, an `I` tag is always preceded by a `B` tag. The $2421$ train samples contained $4695$ unique spans with an average of $1.939$ spans per training instance. Figure \ref{fig:enter-label} outlines the distribution of the number of spans of a given length, and the majority of spans are $\leq 5$ in length. In the train set, each row contained an `id | space-separated token | list of span indices | space-separated gold span label.'
Meanwhile, the $606$ test instances were divided into $182$ public leader board and $424$ privately held instances. 

\section{Task Details}
\textbf{Hosting.} \textit{HateNorm} was hosted as a Kaggle\footnote{\url{https://www.kaggle.com/competitions/hatenorm23}} Competition from 13th July 2023 to 16th August 2023 PST. It received participation from $12$ teams, leading to $72$ submissions (an average of $6$ submissions per team) throughout the competition. As a part of the Kaggle competition, participants were given a sample codebook and a sample `submission.csv,` as outlined in Table \ref{tab:hatenorm_sub}. We required the `id | space-separated predicted label' for the submission file. 

\begin{table}[]
\begin{tabular}{c|c}
\hline
\textbf{Id} & \textbf{Predicted Span List}  \\ \hline
   100  & O O O B I \\
   200 & O O O O \\
   606 & O O O O O B I I I O O O B O \\ \hline
\end{tabular}
\caption{Submission format of test.csv with predictions corresponding to inputs enlisted in Table \ref{tab:bio_samples}.}
\label{tab:hatenorm_sub}
\end{table}

\textbf{Evaluation Metric.} 
Unlike the classification of a single instance that can be adjudged via accuracy or macro-F1, span detection requires evaluating the correct ordering of spans, `B' following a `I' and `O' being the default. To capture this sequential nature of label prediction for individual tokens, we employ the seqeval macro-F1 metric \cite{seqeval}. We hosted the custom metric of seqeval as a script and loaded that to set up the competition so that each incoming submission, by default, gets evaluated via seqeval macro-F1. Further, held 70\% of the test samples were private, based on which the final rankings were revealed after the contest. During the contest, the participants saw their rank compared against the public leaderboard 30\%. Note that a public leader board does not mean test cases are public.

\textbf{Baselines.}
The codebook provided to the participant's finetuned a DistillBERT+FNN setup which reported a extremely low macro-F1 of $0.36$. Meanwhile, the baselines provided by \citet{hatenorm} consisted of BiLSTM+CRF with a macro-F1 of $0.44$, and the best method\footnote{Note: We excluded the Elmo based system from baseline due to reproducibility issues with Elmo on both Tensorflow and Pytorch.} being a SpanBERT \cite{joshi-etal-2020-spanbert}+BiLSTM+CRF system that reported a macro-F1 of $0.58$.

\section{Submitted System}
Table \ref{tab:hatenorm} enlists the top $6$ submissions. Among the participating teams that shared the overview notes, we observed that `FiRC-NLP' employed an ensemble of SpanBERT + CRF with teacher enforcing. When run under lowercase preprocessing, the setup led to the highest macro-F1 of $0.58$. The SpanBERT-based method, `FiRC-NLP,' is also at par with the SpanBERT system of the baseline solution \cite{hatenorm}. Note that owing to a one-to-one mapping of tokens to span tags, we discouraged the users from performing additional preprocessing. Meanwhile, the second-best team `Mohammadmostafa78' with a macro-F1 of $0.52$, overcame the skewness in BIO notations by converting the label space to only BO and employing an XLM-RoBERTa \cite{conneau-etal-2020-unsupervised}+FNN setup. The third highest scoring teams, `CNLP-NITS-PP'  and `IRLab@IITBHU,' have a macro-F1 of $51$, differing only fourth decimal place. However, both employ distinct methods. While the former employs a BERT+BiLSTM+FNN setup, the latter employs contextual embedding (Glove) based BiLSTM+CRF setup akin to the existing baseline. Similar to the observations in our baseline solutions, we observe that BiLSTM and contextual embedding-based solutions perform considerably well. Overall, while Transformer systems either in the form of BERT or SpanBERT help improve the performance, a BiLSTM system trained via CRF is equally viable. We also observe that the proposed systems submissions more or less follow the performance trends of the existing baseline solutions, further corroborating that combining transformer-based systems with CRF and BiLSTM attention mechanisms is the optimal way to detect hateful spans.  

\begin{table}[!t]
\begin{tabular}{c|c|c|c}
\hline
\textbf{Rank (Change in Rank)} & \textbf{Team Name} & macro-F1 & \# Submissions\\ \hline
1 (-)             & FiRC-NLP           & 0.57605 & 8 \\
2 ($\uparrow$3)            & Mohammadmostafa78       & 0.51382 & 2\\
3 ($\downarrow$1)           & CNLP-NITS-PP       & 0.50888 & 22\\
4 ($\downarrow$1)             & IRLab@IITBHU & 0.50861 & 9\\
5 ($\downarrow$1)             & Niranjan Rao & 0.49563 & 4\\
6 ($\downarrow$1) & TextShield & 0.45661 & 5\\ \hline
\end{tabular}
\caption{Top-6 teams based on seqeval macro-F1 on the private leader baord. The ranks are the final rank obtained on private leaderboard, and change in rank is how many positions a team moved up or down in the table when the final ranks were computed on the private board. We also enlist the number of submissions made by the team during the competition.}
\label{tab:hatenorm}
\end{table}

\section{Conclusion}
Despite engaging with malicious content, some online users are adaptable and can be persuaded to change their beliefs through empathy and corrective conduct. Through this task, we aimed to help these users whose social interactions can eventually be nudged to becoming non-hateful. We believe that the proposed systems can be effectively utilized to assist the moderators.

\bibliography{task3_hasoc23}

\end{document}